\documentclass{article}

\usepackage{arxiv}

\usepackage[utf8]{inputenc} % allow utf-8 input
\usepackage[T1]{fontenc}    % use 8-bit T1 fonts
\usepackage{hyperref}       % hyperlinks
\usepackage{url}            % simple URL typesetting
\usepackage{booktabs}       % professional-quality tables
\usepackage{amsfonts}       % blackboard math symbols
\usepackage{nicefrac}       % compact symbols for 1/2, etc.
\usepackage{microtype}      % microtypography
\usepackage{lipsum}		% Can be removed after putting your text content
\usepackage{graphicx}
\usepackage{natbib}
\usepackage{doi}

\usepackage{xcolor}         % colors

\usepackage{amsmath}
\usepackage{amssymb}
\usepackage{xspace}

\usepackage{xcolor}
\usepackage{pifont}
\newcommand{\vm}{{ViewMix}\xspace}
\newcommand{\cmark}{\ding{51}}
\newcommand{\xmark}{\ding{55}}

\title{\vm: Augmentation for Robust Representation in Self-Supervised Learning}

\author{{Arjon Das} \\
	Department of Computer Science\\
	University of Nebraska at Omaha\\
	Omaha, NE 68182 \\
	\texttt{arjondas@unomaha.edu} \\
	\And
	{Xin Zhong} \\
	Department of Computer Science\\
	University of Nebraska at Omaha\\
	Omaha, NE 68182 \\
	\texttt{xzhong@unomaha.edu} \\
}

\renewcommand{\shorttitle}{\textit{ViewMix}}

\hypersetup{
pdftitle={\vm: Augmentation for Robust Representation in Self-Supervised Learning},
pdfsubject={q-bio.NC, q-bio.QM},
pdfauthor={Arjon Das, Xin Zhong},
pdfkeywords={First keyword, Second keyword, More},
}

\begin{document}
\maketitle

% keywords can be removed
% \keywords{Self Supervised Learning \and Augmentation \and Joint Embedding Architecture}

\begin{abstract}
Joint Embedding Architecture-based self-supervised learning methods have attributed the composition of data augmentations as a crucial factor for their strong representation learning capabilities. While regional dropout strategies have proven to guide models to focus on lesser indicative parts of the objects in supervised methods, it hasn't been adopted by self-supervised methods for generating positive pairs. 
This is because the regional dropout methods are not suitable for the input sampling process of the self-supervised methodology. 
Whereas dropping informative pixels from the positive pairs can result in inefficient training, replacing patches of a specific object with a different one can steer the model from maximizing the agreement between different positive pairs. 
Moreover, joint embedding representation learning methods have not made robustness their primary training outcome.
To this end, we propose the \vm augmentation policy, specially designed for self-supervised learning, upon generating different views of the same image, patches are cut and pasted from one view to another. 
By leveraging the different views created by this augmentation strategy, multiple joint embedding-based self-supervised methodologies obtained better localization capability and consistently outperformed their corresponding baseline methods.
We also demonstrate that incorporating \vm augmentation policy promotes robustness of the representations in the state-of-the-art methods.
Furthermore, our experimentation and analysis of compute times suggest that \vm augmentation doesn't introduce any additional overhead compared to other counterparts.
\end{abstract}
\section{Introduction}
\label{sec:intro}

Dependence on a large amount of annotated training data is one of the limiting factors when performing accurate predictive tasks through supervised learning. 
To improve the training efficiency and performance of deep learning models, researchers, on the one hand, explore the enrichment of data, for instance, with augmentation, and on the other hand, investigate unsupervised and semi-supervised learning techniques to reduce data dependency. 

Self-supervised learning (SSL) in computer vision, especially the joint embedding architectures for representation learning, has gained plenty of traction in recent years, with some methods performing as well as the state-of-the-art supervised methods without needing any labeled samples during the pretraining stage. 
While data augmentation techniques have proven to be quite effective in ensuring better training efficacy of supervised learning, they play an even more crucial role in self-supervised pretraining methods for obtaining good representations.

\begin{figure}[h]
    \centering
    \includegraphics[width=0.5\linewidth]{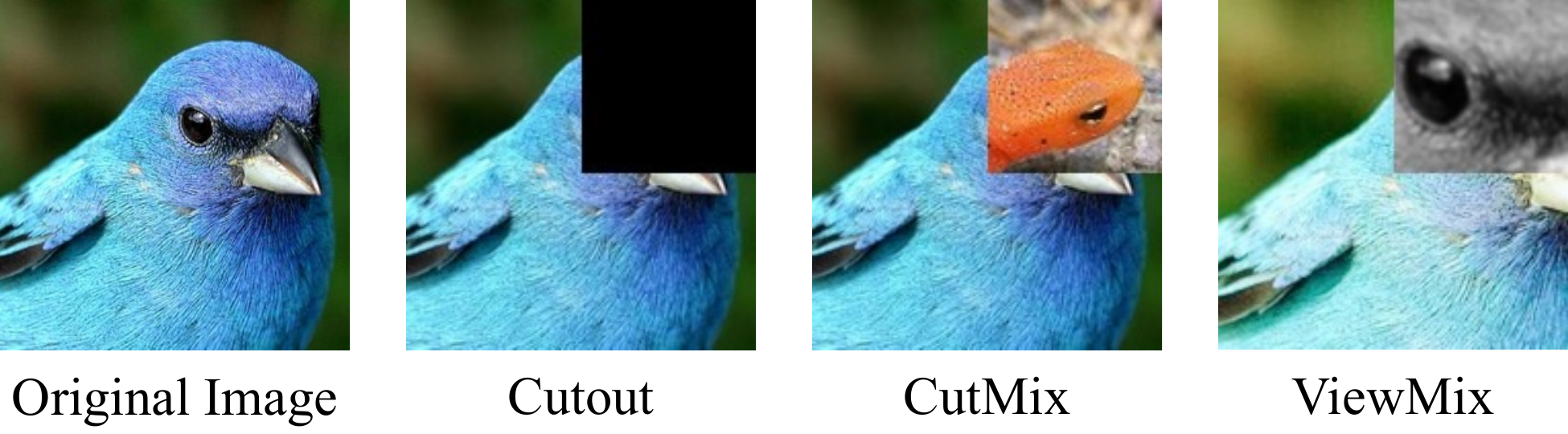}
    \caption{Visualization of Cutout, CutMix and ViewMix augmentation.}
    \label{fig:examples}
\end{figure}

Thorough experimentations~\cite{chen2020simple, bardes2021vicreg, zbontar2021barlow, caron2021emerging, grill2020bootstrap}, over the recent years have shown that augmentation policies like random cropping, random flipping, color distortions, and gaussian blur compounded on top of each other have proven to be very effective means of transformation for the pretext task.
Whether it's contrastive, non-contrastive, redundancy reduction, or asymmetric network methods, these augmentation policies are regularly incorporated in the recent SSL training schemes to produce different views of the same image sample. Here, the term `view'~\cite{chuang2022robust} refers to a transformed image produced after applying multiple data augmentation techniques. Exposing models to different views and optimizing for maximum agreement between the views have demonstrated significant improvement in self-supervised representation learning.

Besides the representative property of the learned features, only a few SSL methods~\cite{chuang2022robust, Yan_2022_CVPR} targeted the robustness against noisy data. 
The robustness refers to the learned representation's insensitivity or invariance to the distortions or augmentations on the inputs. 
Thus, as long as the inputs are the same image, the learned representation should be intact regardless of the augmentations. 
Since the joint embedding architectures experience aggressive forms of image transformations, the models become invariant~\cite{ericsson2022self} to certain distortions. Combining SSL methods has even considerably improved the label noise robustness of supervised methods~\cite{ghosh2021contrastive}. The widespread deployment of deep neural networks in many downstream real-world tasks has made the importance of robustness more appropriate.
% Here robustness refers to retaining models' performance for the out-of-distribution classification of difficult and near-distribution examples.
% Some self-supervised methods have adopted the generative capability of neural networks to perform handcrafted prediction tasks that can lead to useful representations. 
% For instance, grayscale colorization, image super-resolution, image pinpointing, and inverting geometric transformations are some of the tasks that lead to meaningful image representations. 
% But discriminative methods, particularly contrastive methods, outperform these aforementioned self-supervised techniques by a big margin~\xz{citation}.
% Such contrastive methods primarily focus on minimizing differences between similar pairs while maximizing differences between dissimilar pairs.

% Similar to self-supervised learning, robustness, and generalization for unseen shape variations can be obtained by improving localization capability. 
% Deep learning models often face the problem of focusing too much on the small intermediate set of activations or local patches of information. 
% Adopting this narrow outlook eventually provides a weaker representation of general downstream tasks.
% To mitigate this problem, regional dropout strategies\xz{ that ... (briefly define/explain regional dropout)} have been proposed to enhance the performance of the models. 

Similar to self-supervised learning, robustness, and generalization for unseen shape variations can be obtained by better localization capability~\cite{song2019robust}. 
On the contrary, deep learning models often face the problem of focusing too much on the small intermediate set of activations or local patches of information. 
Adopting this narrow outlook provides a weaker representation of general downstream tasks. 
% Although regional dropout (the process of removing informative pixels) strategies~\cite{devries2017improved, zhang2017mixup, yun2019cutmix} are suitable for solving this issue, leading techniques are~\cite{zhang2017mixup, yun2019cutmix} not appropriate for self-supervised learning.
Although regional dropout (the process of removing informative pixels) strategies~\cite{devries2017improved, zhang2017mixup, yun2019cutmix} are suitable for solving this issue for supervised learning, leading techniques are~\cite{zhang2017mixup, yun2019cutmix} not appropriate in self-supervised settings. 
Because part of their optimization relies on utilizing the newly generated mixed labels and SSL techniques don't rely on labels. On top of that such data mixup methods~\cite{zhang2017mixup, yun2019cutmix}, potentially situated in between different classes, don't include any complementary information in the sample in question.
On the other hand, Cutout~\cite{devries2017improved} can lead to training inefficiency due to missing pixels. Moreover, we argue that the current SSL methods' lack of attending local features also results in learning suboptimal feature representations. Due to the correlation of local features with robustness~\cite{song2019robust}, this lack also results in substandard robustness.

In this paper, we propose a novel image augmentation strategy that is particularly designed for self-supervised learning -- \textbf{\vm}, which has three main advantages.
(i) By simply patching one view on top of another to impose a regional dropout and replacement scenario, \vm can be flexibly integrated with different joint embedding learning architectures; 
(ii) Adding \vm along with the standard SimCLR-like image augmentation protocol, we find that the learned representations from multiple state-of-the-art joint embedding learning methods consistently outperform their corresponding baseline (or non-\vm) counterparts on linear evaluations of representative property. 
(iii) We show that the learned representations from adopting \vm with different joint embedding architectures have higher robustness than their corresponding baselines in standard linear classification testing with previously unseen noises.

\section{Related work}
\label{sec:literature}

Unsupervised representation learning frameworks are mostly formulated as generative or discriminative methods. 
Generative methods learn to generate new data instances from input data. 
Learning the data generation process of such models imposes learning the data distribution of the inputs, resulting in intermediate feature maps that can be utilized for input representation. 
For instance, Masked autoencoders~\cite{he2022masked} have demonstrated to learn strong pretext tasks through learning to reconstruct holistic visual concepts. 
Whereas generative methods learn the distribution of data and utilize the intermediate feature maps as image representations, discriminative methods learn to differentiate between types of data instances.
Many self-supervised methodologies follow this goal to maximize agreement between different views of the same image and minimize between different ones. 
Recently, there has been the emergence of non-contrastive methods as well which eliminates the requirements of negative samples necessary for the discriminative methods. 
In this section, we will briefly discuss some of these approaches and analyze the literature concerning the \vm augmentation.

Discriminative methods, particularly contrastive methods have mostly occupied the state-of-the-art chart in self-supervised learning. 
Chen \textit{et al.} proposed the SimCLR~\cite{chen2020simple} method, which is a simple framework for learning representations in a self-supervised manner. 
This framework introduced augmentation-oriented representations learning methodology in addition to the use of projection heads with encoders to establish an excellent learned representation. 
NNCLR~\cite{dwibedi2021little} has extended this instance discrimination task to include non-trivial positives between augmented samples of the same images and among different images. 
These positive samples of near-neighbors are drawn from a support set of image embeddings. 
NNCLR along with other methods, e.g. MoCo~\cite{chen2020improved}, has adopted memory banks in their scheme to maintain the support set of nearest neighbors. 
This increases the complexity of the training schemes and causes a large overhead in memory requirements. 
Additionally, all the contrastive approaches often require comparing each sample with many other samples optimally and the performance varies by the quality of the negative sample pairing. 
This begs the question of whether the negative pairing is essential.

Recently many clustering, asymmetric network learning, and redundancy reduction methods have emerged. 
For instance, DeepCluster~\cite{tian2017deepcluster} bootstraps previous versions of its representations to produce targets for the next one. 
The method clusters data points using current representations which helps it to avoid the usage of negative pairs. 
Dissimilar to DeepCluster, BYOL~\cite{grill2020bootstrap} proposes image representation learning with online and target networks that interact and learn from each other. 
The method also employs a slow-moving average of the online network on the target network to encourage encoding more information within the online projection. 
Zbontar \textit{et al.}~\cite{zbontar2021barlow} proposes Barlow Twins which produces a cross-correlation matrix of the representations close to the identity matrix, forcing strong correlation within each dimension of the representations between the two siamese branches, and decor-relates the pairs of different dimensions. 
But the method relies heavily on batch normalization which prevents collapse when working with only positive samples. 
Non-contrastive methods like VICReg~\cite{bardes2021vicreg}, VIbCReg~\cite{lee2021vibcreg}, VICRegL~\cite{bardes2022vicregl} have also been formulated to answer that question.
Although these different classes of methodologies propose different ideas, they identify and address the sensitivity to choosing the composition of image transformations to result in better image representations. 
Furthermore, Chen \textit{et al.}~\cite{chen2020simple} pointed out that applying cropping in composition with strong color jitters in SSL pretraining has rendered better performance than complex supervised augmentation policies. 

The current state of self-supervised methodologies dominantly uses cropping, horizontal flip, color jitter, Gaussian filter, gray-scaling, and solarization.
An investigation to identify other augmentations is of great interest, which can reinforce the performance of the SSL methodologies. 
In addition, state-of-the-art joint embedding learning mainly focuses on how representative the learned features are, and the invariance or robustness is one of the training methods. 
Through experimentation and analysis, we have proposed a new augmentation policy \vm, which is suitable to the self-supervised learning methods. 
Unlike the previous methodologies, we highlight robustness as one of the primary training outcomes, along with superior image representations.
Experimentation shows that adopting \vm on top of base sets of augmentations during the pretraining of multiple self-supervised methods has consistently resulted in higher linear evaluation accuracy than their base counterpart.
\section{\vm}
\label{sec:method}

This section presents the \vm augmentation in detail.
Section~\ref{sec:method_motivation} discusses the design motivations behind \vm. 
Section~\ref{sec:algorithm} describes the \vm algorithm. 
Section~\ref{sec:flexxibility} talks about the flexibility when using \vm in SSL schemes. 
% (iii) the architecture behind the decoder for image reconstruction, and (iv) the architecture of the projection layer.

\subsection{Motivation}
\label{sec:method_motivation}

Recent research in joint embedding learning has strongly suggested that augmentation policies play a crucial role in obtaining better representations. 
Correct selection of augmentations is critical that using a simple composition of scaling and color distortion during the self-supervised training can guide the model to gain higher linear evaluation accuracy than adopting some of the most sophisticated augmentation policies practiced in supervised techniques. 
On the other hand, regional dropout and replacement strategies have demonstrated their ability to enhance performance in classification tasks by incentivizing feature extractors to focus on less discriminative parts of objects, thus obtaining better object localization capability. 
Furthermore, joint embedding representation learning methods did not control the robustness as their primary training outcome, although some of them applied robustness/invariance as one of the training methods for representative features. 
Motivated by these facts, we have formulated the \vm augmentation. Specifically, \vm initiates a regional dropout and replacement strategy appropriate for the SSL frameworks.

While regional dropout augmentation strategies, for instance, Cutout augmentation, encourage focusing on inconspicuous parts of the object, the dropping of pixels makes the learning process inefficient due to introducing blank information. 
Although masked autoencoders~\cite{he2022masked} work very well by simply applying heavy information dropout, they only work with ViT-based~\cite{dosovitskiy2020image} architectures. 
On the contrary, CutMix~\cite{yun2019cutmix} augmentation mitigates the learning inefficiency of Cutout augmentation by filling in blank pixels of the training sample with a patch from another object sample. 
Although the method has proven effective for supervised methods, such augmentation does not fit well with joint embedding learning, where maximizing agreement between different views of the same image is the goal. 
Since CutMix replaces image regions with a patch of another image, incorporating it in joint embedding learning introduces different views referring to two different classes of objects rather than from a single one. 
In such a scenario, the objective of the SSL training doesn't correlate with the augmentation. 
Later in the experiment section, we will observe that CutMix establishes a more impaired pretraining condition for SSL.

We designed \vm augmentation specifically to address the issues mentioned earlier by Cutout and CutMix under the SSL criteria. 
The augmentation is inspired by Cutout and CutMix and is suitable for joint embedding learning architectures. Unlike CutMix, which replaces the region of the training sample with a patch of a different image of a different class, \vm takes two different views of the same image, replacing the region of one of the views with a patch from the other. 
The views are generated from the standard SSL transformations of the original image. 
The key differences between Cutout, CutMix, and \vm are summarized in table~\ref{tab:feat_summary}.

\begin{table}[]
  \centering
  \caption{Summarization of Cutout, CutMix, and ViewMix.}
  \resizebox{0.45\textwidth}{!}{
  \begin{tabular}{@{}lccc@{}}
    \toprule
     &Cutout &CutMix &ViewMix \\
    \midrule
    Regional dropout & \cmark & \cmark & \cmark \\
    Full image utilization & \xmark & \cmark & \cmark \\
    Suitable for SSL & \cmark & \xmark & \cmark \\
    \bottomrule
  \end{tabular}
  }
  \label{tab:feat_summary}
\end{table}

\begin{figure*}[ht!]
    \centering
    \includegraphics[width=0.9\linewidth]{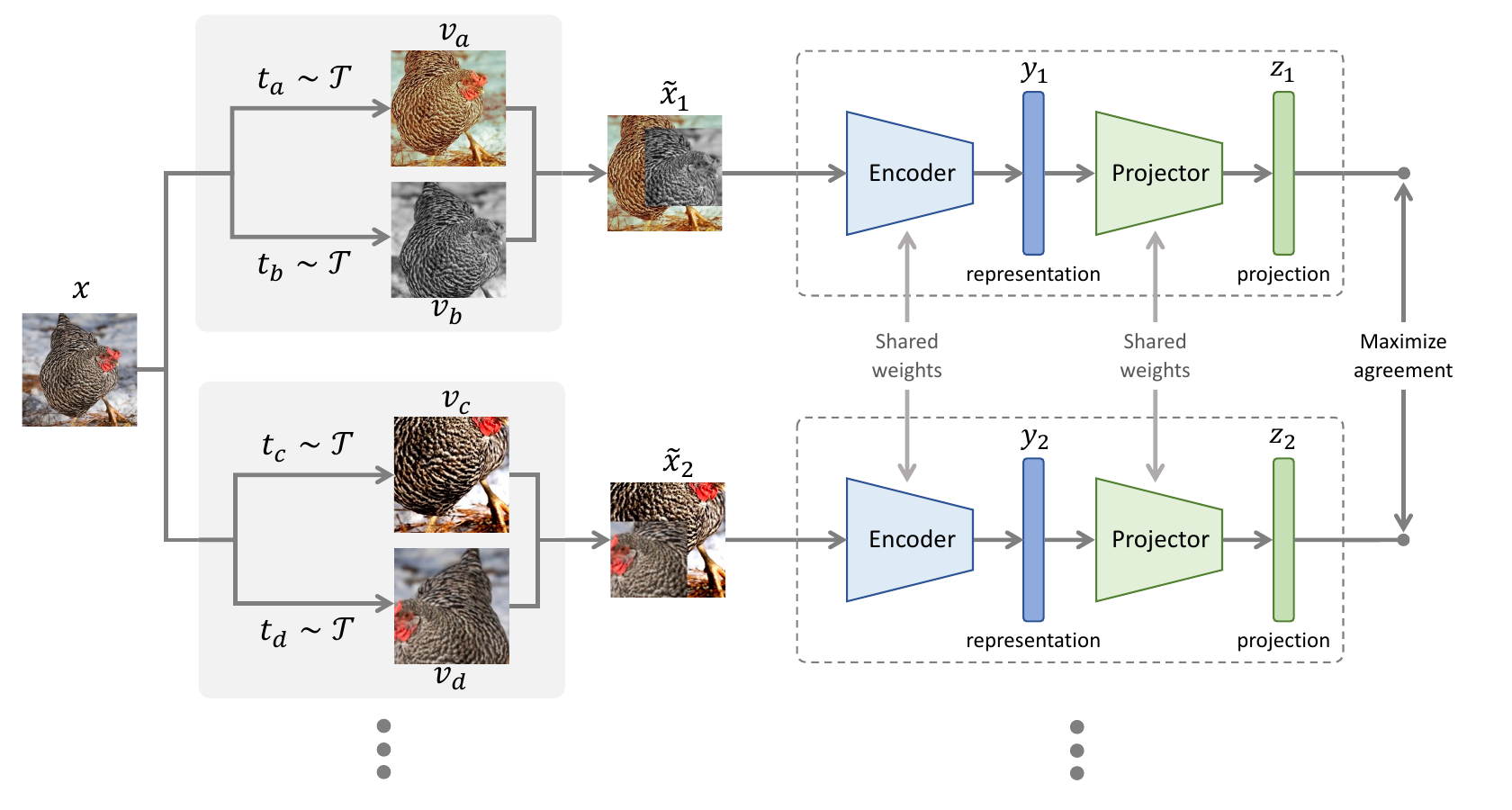}
    \caption{Summary of \vm augmentation.}
    \label{fig:viewmix}
\end{figure*}

Fig:~\ref{fig:viewmix} illustrates the \vm augmentation process. 
First, the original image is processed through two transformations of the same distribution to generate two unique views, $A$ and $B$. 
Then, we replace the region of $A$ with a random patch sampled from view $B$. 
After patching, a new view $A'$ is formulated, which continues to the SSL pretraining process.

\subsection{Algorithm}
\label{sec:algorithm}

\vm augmentation is designed to leverage the transformation stage of the recent joint embedding learning schemes. 
During pretraining, for a given image $x \in \mathbb{R}^{W \times H \times 3}$, sampled from dataset $\mathcal{D}$, $t_1, t_2, ... t_n$ transformations are applied to produce $n$ different views $v_1 = t_1(x), v_2 = t_2(x), ..., v_n = t_n(x)$. Here, $t_1, t_2, ..., t_n$ are sampled from a distribution $\mathcal{T}$, $n > 1$, and $W$ and $H$ is the width and height of each input image. 
In most joint embedding learning processes, each of these transformations is a predominantly random crop of the sample $x$ followed by color distortions. 
The goal of the \vm augmentation is to generate a new training sample $\Tilde{x}$ by masking two different views produced by any two transformations $t_a$ and $t_b$ from distribution $\mathcal{T}$ with mask $\mathbf{M}$. 
Here, $\mathbf{M} \in \{0, 1\}^{W \times H}$ is a binary mask used to indicate the pixel information to be swapped by the ones of a different view. 
The sample $\Tilde{x}$ is then used to continue the joint embedding training. We define the augmentation process as follows:

\begin{equation}
    v_a = t_a(x),
\end{equation}

\begin{equation}
    v_b = t_b(x),
\end{equation}

\begin{equation}
    \Tilde{x} = \mathbf{M} \odot v_a + (1 - \mathbf{M}) \odot v_b.
    \label{eq:masking}
\end{equation}

\noindent 
The masking region of M is filled by 0 and the remaining by 1. 
Consequently, the region with 0's replaces the pixel information with another view's information while the region with 1's is kept intact. 
The masking region is defined by a bounding box containing center point coordinates $(b_x, b_y)$ and $b_w$ and $b_h$ as the width and height of the bounding box, respectively. 
Given a view of width $W$ and height $H$, we obtain $b_x$ and $b_y$ by uniformly sampling from the range $[0, W]$ and $[0, H]$. 
The width $b_w = \lambda \times W$ and height $b_h = \lambda \times H$ of the bounding box preserve the aspect ratio of the original view. 
$\lambda$ is a fraction uniformly sampled from the range $[r_{min}, r_{max}]$, where $0 < r_{min} < r_{max} < 1$.
$r_{min}$ and $r_{max}$ can be exposed as hyperparameters to guide the area of the randomly replaced view per augmentation.
% \cite{devries2017improved} has also suggested that the shape of the mask $\mathbf{M}$ contribute to any significant changes in the learned representation, rather the size of the mask if more crucial.

\subsection{Flexibility}
\label{sec:flexxibility}

% \textbf{Why is \vm flexible with SSL?} 
To produce a single sample of \vm augmented image; first, we need to run two transformations to generate two different views of the image. 
The transformation pipeline is inspired by the standard SimCLR framework and each of these transformations $t_i$ (sampled from the distribution $\mathcal{T}$) consists of random cropping followed by multiple color distortion operations and random horizontal flips. 
It seems like the \vm augmentation initiates a lot of computational overhead. 
However, since most joint embedding learning architecture requires generating two or more views from a single sample, we can use those views directly in our augmentation to formulate the final augmented input. 
This implementation trick prevents \vm from initiating any additional transformations; hence there is no additional computational overhead. 
Consequently, \vm augmentation is flexibly integrated with most joint embedding learning schemes, whether it is a contrastive, asymmetric network, redundancy reduction, or non-contrastive method.
\section{Experiments and Analysis}
\label{sec:experiment}

In this section, we evaluate the representations learned by multiple self-supervised method-based pretraining when applied with \vm. 
The evaluation focuses on \vm's efficacy in improving the localizability and generalizability of representations obtained from these pre-trainings. 
The evaluation process is three-fold. 
First, we evaluate the learned representations with linear classification applied to five different SSL methods. 
We demonstrate that the addition of the \vm improves the linear classification accuracy across five different popular SSL methods, namely SimCLR, VICReg, BYOL, Barlow Twins, and VIbCReg, compared to their corresponding base composition of transformations. 
Second, we evaluate the robustness of the representations obtained from these pre-trainings. 
In that manner, we show that the \vm augmentation policy consistently improves frozen linear classification accuracy even when introduced with previously unseen augmentations, indicating robustness in the representations. 
Third, to further evaluate the effectiveness of downstream tasks, we finetuned the model for multiple few-shot recognition tasks and a segmentation task.
Due to computational constraints, we kept our pretraining limited mostly to the CIFAR10 dataset. For segmentation evaluation, we pretrained on ImageNet dataset with SimCLR, VICReg, and Barlow Twins method paired with different augmentation strategies.
We also visualize and compare the Class Activation Mappings (CAM) of different augmentations and analyze their effects.
Finally, we analyze if there is any computational overhead introduced by our proposed method.

\subsection{Linear Evaluation}
\label{sec:linear_eval}

\paragraph{Evaluation Method.} 
For evaluation, we have selected five joint embedding-based SSL methods: SimCLR, VICReg, VIbCReg, BYOL, and Barlow Twins. 
ResNet-18 architecture is used as the backbone for all the methods. 
All the SSL methods have the same composition of transformations. 
Specifically, the SSL frameworks include the five standard image transformations randomly applied and compounded on top of each other.
These transformations are Cropping + Rescaling, Color Jitter, Grayscale, Gaussian, and Solarization. 
We pre-train each of the SSL methods with and without the ViewMix augmentation and freeze the weights. 
We remove the projection layer and attach a linear layer (initialized with random weights) with the frozen weights of the backbone and train for classification. 
The corresponding validation accuracy of the finally obtained classifier is linear evaluation accuracy. 
In all experiments, we train the linear classifier for 100 epochs using the Adam optimizer with labeled training set data. 
The performance comparison across all the SSL methods with and without the \vm augmentation along with the base standard augmentations helps portray the superiority of the learned representations on downstream computer vision tasks.

\paragraph{Image Transformation Details.}
SSL frameworks depend heavily on image transformations to produce different views of the same object. 
The following are the brief details of augmentations that are applied during the training: (i) Image cropping with random sizes from 75\% to 100\% of the original area; (ii) Random horizontal flip of the images with 0.5 probability; (iii) Random Color Jittering with a probability of 0.8; (iv) Random Gaussian Filter with a probability of 0.2; (v) Gray-scaling with a probability of 0.2; (vi) Solarization with a probability of 0.2; and (vii) \vm with a probability of 0.33 (if applied).

Some of the transformation intensities have minor variations depending on which SSL methods it is applied to. But the random application of ViewMix is kept to 33\% on all the methods and the area is randomly selected between 30\% to 60\% of the area.

\paragraph{Analysis of Results.}
Table~\ref{tab:linear_eval} illustrates linear evaluation accuracy of different self-supervised learning methods which represent \vm augmentation policy against the base transformations. 
We can observe that the addition of \vm increases the linear evaluation accuracy in all cases. 
For some methods, significant accuracy gain upon the addition of \vm can be obtained. 
For instance, VIbCReg improves by $\textbf{+2.34\%}$ and VICReg improves by $\textbf{+1.74\%}$ when the pretraining has \vm augmentation policy along with the SimCLR-style baseline augmentations.
Table~\ref{tab:imagenet} demonstrates the linear evaluation accuracy of different SSL methods on the Imagenet dataset. 
Although \vm's top-1 accuracy of the linear classification is slightly lower, it provides a better representation of the semantic segmentation on the Oxford-IIIT Pet dataset. 
We notice that \vm continuously increases the performance as we increase the size of the dataset.
As we are approaching the upper limits on what our current training hardware can allow, we conjecture that if we train with a much larger number of examples and expand the parameter search, \vm could further improve the classification accuracy and provide better regularization. 

\begin{table}[]
\caption{Linear Evaluation accuracy of SimCLR, VICReg, BYOL, Barlow, VIbCReg with Baseline, Cutout, \vm and Cutout+\vm augmentation, after 1000 epochs and CutMix augmentation after 200 epochs of pretraining on CIFAR-10 dataset.}
\label{tab:linear_eval}
\resizebox{\textwidth}{!}{%
\begin{tabular}{lccccccccccccccc}
\toprule
ResNet-18 &
  Epochs &
  \multicolumn{2}{c}{SimCLR} &
   &
  \multicolumn{2}{c}{VICReg} &
   &
  \multicolumn{2}{c}{BYOL} &
   &
  \multicolumn{2}{c}{Barlow} &
   &
  \multicolumn{2}{c}{VIbCReg} \\ \cmidrule{3-4} \cmidrule{6-7} \cmidrule{9-10} \cmidrule{12-13} \cmidrule{15-16} 
Num of Params: 11.17M &      & Top-1 & Top-5 &  & Top-1 & Top-5 &  & Top-1 & Top-5 &  & Top-1 & Top-5 &  & Top-1 & Top-5 \\ \midrule
Baseline              & 1000 & 90.24 & 99.72 &  & 90.52 & 99.64 &  & 91.98 & 99.81 &  & 91.43 & 99.78 &  & 88.55 & 99.63 \\
+ Cutout              & 1000 & 91.07 & 99.80 &  & 91.86 & 99.76 &  & 92.39 & 99.85 &  & 91.78 & 99.80 &  & 89.62 & 99.69 \\
+ ViewMix             & 1000 & 91.44 & 99.79 &  & 92.26 & 99.78 &  & 92.55 & 99.87 &  & 91.80 & 99.85 &  & 90.89 & 99.81 \\
+ (Cutout+ViewMix)    & 1000 & 91.32 & 99.72 &  & 91.66 & 99.77 &  & 92.45 & 99.78 &  & 90.58 & 99.74 &  & 90.54 & 99.76 \\ \midrule
Baseline              & 200  & 85.18 & 99.50 &  & 89.37 & 99.60 &  & 86.07 & 99.6  &  & 87.47 & 99.62 &  & 85.97 & 99.48 \\
+ CutMix              & 200  & 67.03 & 97.39 &  & 74.74 & 98.18 &  & 71.11 & 97.6  &  & 71.57 & 97.67 &  & N/A   & N/A   \\
+ ViewMix             & 200  & 85.72 & 99.50 &  & 90.17 & 99.74 &  & 84.15 & 99.48 &  & 86.08 & 99.66 &  & 87.94 & 99.71 \\ \bottomrule
\end{tabular}%
}
\end{table}

\begin{table}[]
\centering
\caption{Linear Evaluation (Partial ImageNet) and Segmentation Finetuning (Oxford-IIIT Pet) results}
\label{tab:imagenet}
\resizebox{0.7\textwidth}{!}{
\begin{tabular}{llcccc}
\toprule
ResNet-18             &  & \multicolumn{2}{c}{Linear Classification} &  & Semantic Segmentation      \\
Num of Params: 11.17M &  & \multicolumn{2}{c}{ImageNet}           &  & Oxford-IIIT Pet (Finetune) \\ \cmidrule{3-4} \cmidrule{6-6} 
                      &  & Top-1 (\%)          & Top-5 (\%)          &  & IoU                        \\ \midrule
SimCLR                &  & 74.62               & 93.26               &  & 0.8959                     \\
+ Cutout              &  & 74.84               & 93.46               &  & 0.8952                     \\
+ ViewMix             &  & 74.52               & 93.44               &  & 0.8989                     \\ \midrule
VICReg                &  & 75.94               & 93.48               &  & 0.8928                     \\
+ Cutout              &  & 75.74               & 93.26               &  & 0.8906                     \\
+ ViewMix             &  & 75.40               & 93.38               &  & 0.8919                     \\ \midrule
Barlow                &  & 76.70               & 93.60               &  & 0.8928                     \\
+ Cutout              &  & 76.26               & 93.54               &  & 0.8940                     \\
+ ViewMix             &  & 76.02               & 93.98               &  & 0.8943                     \\ \bottomrule
\end{tabular}
}
\end{table}

\paragraph{\vm and CutMix.}
The \vm augmentation has some of the resemblance and characteristics of CutMix, yet when applied to Joint Embedding SSL methods, they fall apart. 
Firstly, apart from the better localization effect, CutMix offers mixed labels, which is unnecessary in the case of SSL. 
Secondly, joint embedding learnings are focused on maximizing agreement between different views of an object. 
Since CutMix patches another object class onto the input image, the optimization toward view agreement becomes confusing because the model is experiencing samples containing a patch of a different object class. 
Consequently, CutMix results in suboptimal self-supervised learning. We can observe these behaviors in Table~\ref{tab:linear_eval}, comparing the linear evaluation accuracy of \vm and CutMix on multiple state-of-the-art joint embedding-based SSL methods. 
In 200 epoch pretraining, CutMix augmentation with different SSL methods has consistently performed worse than their corresponding baselines, and \vm has shown improvement upon the baseline. 
In the case of VIbCReg, in all the training iterations, the pretraining failed to finish with CutMix.

\paragraph{Comparison with VICRegL.}
VICRegL~\cite{bardes2022vicregl} is a recently proposed self-supervised technique to learn features at a global and local scale. 
It utilizes the VICReg~\cite{bardes2021vicreg} criterion on the pair of feature vectors for maximizing agreement between views. 
VICRegL is an improvement of VICReg. 
We compare the linear evaluation accuracy between the baseline VICRegL and VICReg plus \vm on the CIFAR-10~\cite{krizhevsky2010convolutional} dataset. 
We have selected ResNet-18 as the backbone and chosen $\alpha= 0.75$ for training VICRegL with $256$ batch size. 
The models are trained for $1,000$ epochs. 
We observe that, given these experiment configurations, \vm plus VICReg can outperform the linear evaluation top-1 accuracy of the newly proposed improvement of VICReg (VICRegL).

\begin{table}[h!]
  \centering
  \caption{Linear Evaluation results of VICReg+\vm and VICRegL after $1000$ epochs training.}
  \resizebox{0.4\textwidth}{!}{
  \begin{tabular}{@{}lcc@{}}
    \toprule
     ResNet-18 & Top-1 & Top-5 \\
     Num of Params: 11.17 M & Acc (\%) & Acc (\%) \\
    \midrule
    VICRegL ($\alpha = 0.75$) & 89.02 & 100.00 \\
    VICReg + \vm & \textbf{92.26} & 99.78 \\
    \bottomrule
  \end{tabular}
  }
  \label{tab:linear_eval_sup}
\end{table}

\subsection{Robustness Evaluation}

This subsection discusses our evaluation of the robustness. 
The robustness effect of \vm is compared against Cutout. 
Also, comparing the performance of a base model with and without the \vm is an ablation study highlighting the importance of \vm in achieving robustness.

\paragraph{Evaluation Method.}
The goal of the robustness evaluation is to analyze how well the obtained representations represent their corresponding classes after introducing previously unseen transformations. 
This evaluation process is similar to linear evaluation with just one exception. 
In linear evaluation, after training the classifier with a labeled dataset we run validation on images without any augmentations, or simply putting, inputs are from the same distribution. 
But in robustness evaluation, after the classifier training, we run the validation with previously unseen transformations. Hence the validation is conducted with samples of different data distributions. 
For our evaluation, we have selected Rotation, Rot90, Perspective, and Translation transformation. 
This means after training the classifier with the original unaugmented image dataset, we create four different validation sets which only apply these four augmentations. 
The validation accuracy from those datasets represents our robustness evaluation metric. 
We selected rotation, rot90, perspective, and translation augmentation because these augmentation policies are significantly different from the base augmentations applied during the SSL pretraining, bolstering the fact that augmentations of such nature have not been experienced by the backbone before.

\paragraph{Analysis of Results.}
For fair experimentation, each SSL scheme is trained for 1000 epochs with implementations that result in deterministic transformations (meaning the transformations are pre-generated and cached for later use). 
We've conducted the robustness evaluation on base SSL methods, SSL methods pre-trained with Cutout, and SSL methods pre-trained with \vm. 
For each training run, we logged the resulting validation accuracy in Table~\ref{tab:robustness}.

\begin{table}[]
\caption{Robustness comparison of SSL methods' representations with Baseline, Cutout, ViewMix pretaining, with previously unseen augmentations (Rotation, Rotation-90, Perspective, Translation) during test set inference.}
\label{tab:robustness}
\resizebox{\textwidth}{!}{%
\begin{tabular}{lccccccccccccccc}
\toprule
 &
  \multicolumn{3}{c}{Rotation} &
   &
  \multicolumn{3}{c}{Rotation 90} &
   &
  \multicolumn{3}{c}{Perspective} &
   &
  \multicolumn{3}{c}{Translation} \\ \cmidrule{2-4} \cmidrule{6-8} \cmidrule{10-12} \cmidrule{14-16} 
 &
  Base &
  Cutout &
  ViewMix &
  \multicolumn{1}{l}{} &
  Base &
  Cutout &
  ViewMix &
  \multicolumn{1}{l}{} &
  Base &
  Cutout &
  ViewMix &
  \multicolumn{1}{l}{} &
  Base &
  Cutout &
  \multicolumn{1}{l}{ViewMix} \\ \midrule
SimCLR  & 66.77 & 67.16 & 73.40 &  & 35.25 & 34.30 & 37.18 &  & 74.82 & 75.67 & 77.31 &  & 84.87 & 88.71 & 88.83 \\
VICReg  & 72.64 & 73.24 & 77.61 &  & 36.10 & 36.60 & 40.69 &  & 75.92 & 76.98 & 78.19 &  & 87.47 & 90.03 & 90.03 \\
BYOL    & 74.80 & 74.78 & 77.66 &  & 39.06 & 38.52 & 42.15 &  & 76.55 & 76.83 & 77.92 &  & 89.12 & 90.32 & 90.20 \\
Barlow  & 74.01 & 73.35 & 78.93 &  & 36.70 & 38.15 & 41.53 &  & 77.85 & 78.58 & 79.31 &  & 87.73 & 89.60 & 88.87 \\
VIbCReg & 70.13 & 72.56 & 76.02 &  & 36.09 & 36.60 & 39.67 &  & 75.01 & 76.80 & 77.84 &  & 84.80 & 87.06 & 88.03 \\ \bottomrule
\end{tabular}%
}
\end{table}

Results in each row represent the linear evaluation accuracy of each pretrained model with different SSL frameworks. 
We can see that most of the SSL methods pretrained with \vm augmentation has resulted in higher linear classification accuracy, in previously unseen transformations, namely, Rotation, Rotation 90-degree, Perspective and Translation. 
For instance, when SimCLR is pretrained with \vm, adding rotation augmentation in the linear evaluation validation has \textbf{6.63\%} higher accuracy than the base SimCLR. 
These results highlight the fact that when trained with \vm, feature extractors are less sensitive to the distortions. 
This showcases \vm's superiority on obtaining better robustness with previously unseen image perturbations. 
We attribute this higher robustness to the better localization capability of the SSL methods from the \vm augmentation.
% From the rest of the columns, we can observe that \vm augmentation provides better linear classification accuracy even for previously unseen augmentations compared to the other augmentation frameworks, demonstrating the higher robustness of the SSL models due to the adoption of this augmentation.

\subsection{Transfer Learning of Pretrained Models}

This subsection discusses how models trained with baseline, Cutout, and ViewMix compare when the learned representations are used for transfer learning. 

\paragraph{Evaluation Method.}
Self-supervised learning is aimed toward better feature representation which can be later used for transfer learning on downstream tasks. 
So we examine whether ViewMix augmentation on different SSL methods results in better performance in downstream tasks compared to their baseline and Cutout counterparts. 
For evaluation, we utilized our ResNet18 pretrained weights for transfer learning on Few Shot and Segmentation task. 
More specifically, for Few Shot recognition, we finetuned CIFAR10 pretrained models with Prototypical Network~\cite{snell2017prototypical} across six datasets, namely CIFARFS~\cite{bertinetto2018meta}, Fewshot-CIFAR100~\cite{oreshkin2018tadam}, Caltech-UCSD Birds (CUB)~\cite{wah2011caltech}, Omniglot~\cite{lake2019omniglot}, Double MNIST~\cite{mulitdigitmnist}, Triple MNIST~\cite{mulitdigitmnist}. We consider a 5-way 5-shot transfer, and the test shot always has 32 images per class except for Omniglot, it's 20. For the Segmentation task, we are using ImageNet pretrained weights from VICReg, SimCLR, and Barlow Twins SSL methods. We used Featured Pyramid Network (FPN)~\cite{lin2017feature} with ResNet18 backbone. The segmentation evaluation is based on the Oxford-IIIT Pet Dataset~\cite{parkhi12a} and reports the Intersection over Union (IoU) metric.

\paragraph{Analysis of Results.}
For both types of experiments, we finetuned the existing models with pretrained weights. 
Table \ref{tab:transfer} demonstrates the performance accuracy of different methods with baseline, Cutout and ViewMix strategy. Table~\ref{tab:imagenet} shows the segmentation finetuning efficiency. 
% The weights are obtained via ImageNet100 training which being compute-intensive, has to be limited to SimCLR, VICReg, and Barlow Twins method. 
% Although ImageNet pretraining with \vm results show lower linear evaluation accuracy, 
Finetuning the weights for the Segmentation task results in better IoU over SimCLR and Barlow Twins-based baseline and Cutout strategy. 
In table~\ref{tab:transfer}, we can observe that when finetuning for 5-way 5-shot learning, encoders with \vm-based weights consistently outperform their corresponding baseline and Cutout counterparts in most of the few shot datasets, except Omniglot. %% Talk about domain difference

\begin{table}[]
\centering
\caption{5-way 5-shot training results to evaluate representation transferability}
\label{tab:transfer}
\resizebox{0.8\textwidth}{!}{%
\begin{tabular}{llcccccc}
\toprule
Few Shot Recognition &  & CIFARFS & CIFAR100 & CUB   & Omniglot & Double MNIST & Triple MNIST \\ \midrule
SimCLR               &  & 0.820   & 0.850    & 0.800 & 0.987    & 0.921        & 0.954        \\
+ Cutout             &  & 0.813   & 0.855    & 0.813 & 0.987    & 0.912        & 0.950        \\
+ ViewMix            &  & 0.840   & 0.864    & 0.832 & 0.982    & 0.923        & 0.962        \\ \midrule
VICReg               &  & 0.817   & 0.839    & 0.810 & 0.984    & 0.882        & 0.935        \\
+ Cutout             &  & 0.830   & 0.856    & 0.826 & 0.984    & 0.884        & 0.939        \\
+ ViewMix            &  & 0.840   & 0.860    & 0.847 & 0.980    & 0.898        & 0.941        \\ \midrule
Barlow               &  & 0.842   & 0.864    & 0.830 & 0.989    & 0.906        & 0.944        \\
+ Cutout             &  & 0.838   & 0.862    & 0.829 & 0.988    & 0.896        & 0.946        \\
+ ViewMix            &  & 0.856   & 0.873    & 0.860 & 0.985    & 0.900        & 0.949        \\ \midrule
BYOL                 &  & 0.859   & 0.879    & 0.866 & 0.983    & 0.918        & 0.954        \\
+ Cutout             &  & 0.861   & 0.881    & 0.866 & 0.984    & 0.916        & 0.952        \\
+ ViewMix            &  & 0.862   & 0.884    & 0.880 & 0.981    & 0.919        & 0.959        \\ \midrule
VIbCReg              &  & 0.829   & 0.865    & 0.810 & 0.990    & 0.909        & 0.947        \\
+ Cutout             &  & 0.832   & 0.864    & 0.819 & 0.990    & 0.918        & 0.953        \\
+ ViewMix            &  & 0.856   & 0.877    & 0.856 & 0.988    & 0.915        & 0.953        \\ \bottomrule
\end{tabular}%
}
\end{table}

\subsection{CAM Analysis}

\begin{figure}[h]
    \centering
    \includegraphics[width=\linewidth]{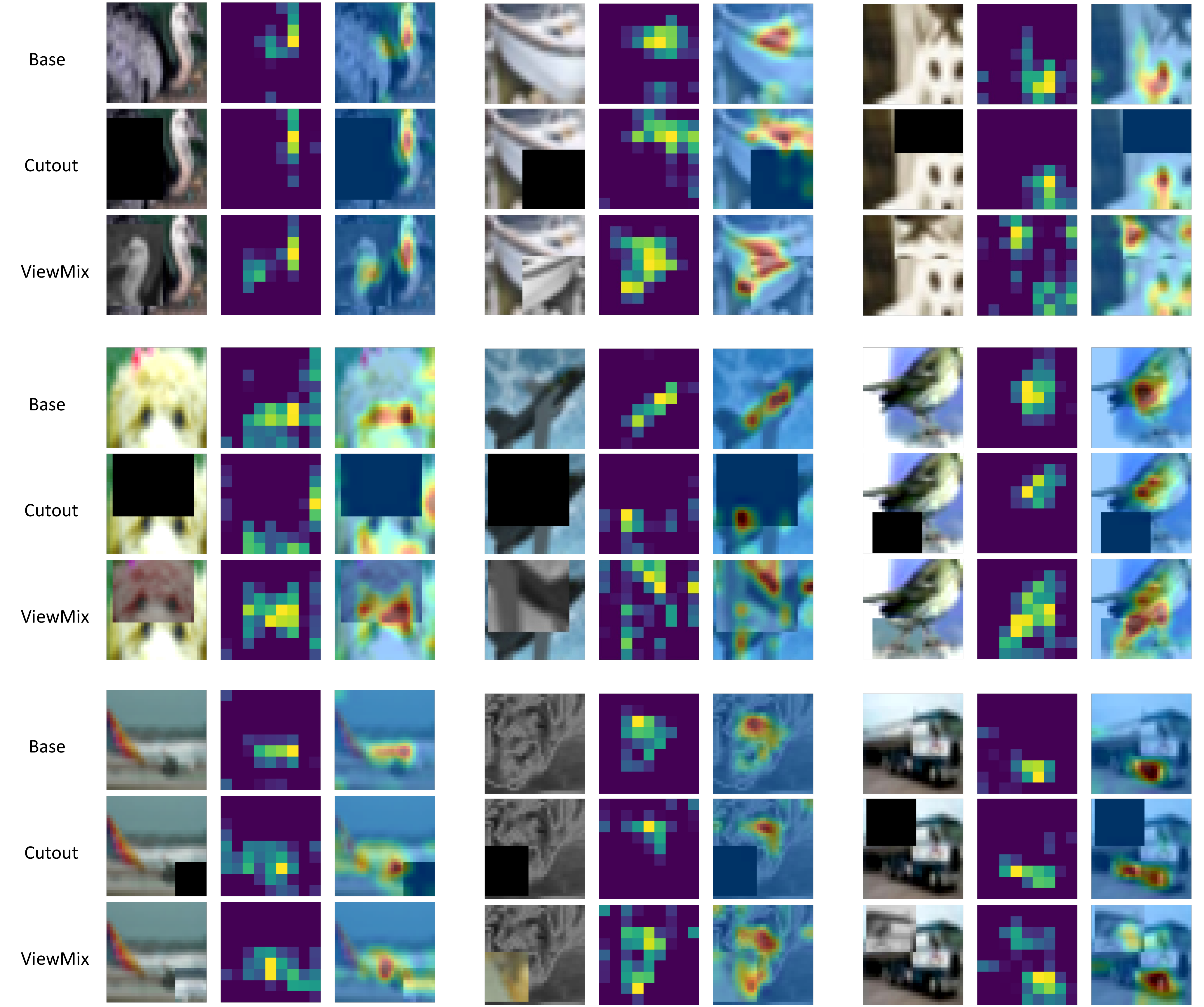}
    % \caption{Class Activation Mapping (CAM) of a CIFAR-10 `deer' class sample. From top to bottom, we have the augmented sample, (only) class activation mapping, and overlaid CAM on the input sample. From left to right we have the baseline SimCLR-style transformation, Cutout and \vm.}
    \caption{Class Activation Mapping (CAM) of multiple CIFAR-10 samples. From left to right in each block, we have the augmented sample, (only) class activation mapping, and overlaid CAM on the input sample. From top to bottom we have the baseline SimCLR-style transformation, Cutout and \vm.}
    \label{fig:cam}
\end{figure}

Regional dropout techniques encourage classifiers to emphasize less obvious features of the image. 
From previous discussions, we have seen that Cutout augmentation brings in training inefficiency and, in the case of SSL, CutMix results in inferior representation learning. 
\vm facilitates dropout with full usage of the image and improves the localization capability by situating a partial view on top of the view in consideration. 
The class activation mappings shown in Fig~\ref{fig:cam} visually demonstrate different classifiers' behavior for the same images. 
The figure consists of multiple blocks of images and each ($3 \times 3$) block from top to bottom upholds three scenarios. Namely, the class activation heatmaps for the classifier pretrained with baseline SimCLR-inspired transformation, Cutout, and ViewMix, respectively. 
From left to right of each ($3 \times 3$) block, we have the input images from CIFAR-10, the class activation map, and the overlaid heatmap visualization of the corresponding input. 
% The figure from left to right upholds three scenarios, baseline SimCLR transformation, Cutout, and ViewMix. 
% From top to bottom, we have the input image ('deer' class from CIFAR-10), the class activation map, and the overlaid heatmap visualization of the corresponding class. 
% To obtain the classifier, we used the ResNet-18 model pre-trained with the base VICReg SSL method on the CIFAR-10 dataset for 1000 epochs and finetuned with a linear layer.

In these examples, encompassing all three augmentation scenarios, we are investigating the model's behavior in terms of image localization and pixel information utilization during classification.
We have observed that when using the base SimCLR transformations, the model predominantly focuses on the prominent features of the object class. 
However, when employing Cutout augmentation, the heatmap extends to non-salient areas, suggesting improved localization. It's also noteworthy that the model does not effectively utilize blank spaces for detection, which can result in training inefficiency.
When utilizing \vm, we notice that the CAM activation heatmap expands even further to encompass the partial view overlaid on top of the original view. This expansion not only enhances localization but also allows the \vm-based model to effectively utilize pixel information from the partial views, thereby contributing to improved training efficiency.
% With \vm, we observe cam activation from both the partial and original view of the input sample. Hence, \vm is an improvement on top of the base SimCLR style transformation by promoting better localization. Additionally, \vm promotes full usage of the pixel information presented to the feature extractor, as opposed to Cutout augmentation.

\subsection{Computational Overhead Time}

The ViewMix augmentation requires generating multiple views of the same image for overlaying one view to another. Consequently, the method apparently suggests a higher computation time for each round of transformation to complete. In section \ref{sec:flexxibility}, we briefly explained that ViewMix augmentation leverages joint embedding learning architectures' multi-view transformation scheme, which prevents it from additional computational overhead. To back that statement, we benchmarked the computation time for the full SimCLR Baseline, ViewMix, Cutout, and CutMix transformation pipeline. For a fair comparison, the transformations were conducted on the exact same hardware specifications for two different image resolutions as shown in Table~\ref{tab:overhead}. From the table, we can observe that \vm execution time is a bit higher than the base SimCLR augmentation pipeline but takes less time than the Cutout and CutMix pipeline. Since other SSL methods also employ SimCLR-style transformations, the analysis is also applicable to those methods.

\begin{table}[]
\centering
\caption{Execution Time of full augmentation pipeline adopted with different strategies. The experiment only computes the transformation round, no forward or backward pass is included.}
\label{tab:overhead}
\resizebox{0.8\textwidth}{!}{
\begin{tabular}{lccccccc}
\toprule
Image Resolution &  & Batch Size & Steps & \multicolumn{4}{c}{Augmentation Pipeline Execution Time (seconds)} \\ \cmidrule{5-8} 
                 &  &            &       & Baseline        & ViewMix        & Cutout         & CutMix         \\ \midrule
$32~\times~32$          &  & 128        & 1000  & 125.827         & 137.194        & 143.568        & 135.319       \\
$224~\times~224$        &  & 128        & 200   & 380.159         & 399.720        & 402.070        & 424.134        \\ \bottomrule
\end{tabular}
}
\end{table}
\section{Conclusion}
\label{sec:conclusion}

This paper introduces \vm, a simple augmentation for joint embedding-based self-supervised image representation learning that promotes localization with regional dropout and replacement of view. 
\vm is straightforward to implement and can be flexibly integrated with SSL pretraining methods. 
With proper reuse of the different views from the SSL transformations, the augmentation adds no computational overhead. 
On ResNet-18-based CIFAR-10 linear evaluation, applying ViewMix with SimCLR, VICReg, BYOL, Barlow Twins, and VIbCReg improves the performance of the baseline by \textbf{1.20\%}, \textbf{1.74\%}, 0.57\%, 0.37\%, \textbf{2.34\%}, respectively. 
Furthermore, we have shown that simply integrating \vm with these methods has resulted in image representations that are more robust by significant margins to previously unseen distortions than the baseline methods.
Finally, this work highlights the potential of augmentations on the self-supervised representation learning process and, how applying specially designed augmentations, without bringing changes to the architecture or learning scheme, results in better representations. 
% We performed our training on several datasets with an increasing number of examples.
% We notice that as we continually increase the size of the dataset, the performance improves.
% As we are approaching the upper limits on what our current training hardware can allow, we conjecture that training with a much larger number of examples could improve the performance even further. 

\bibliographystyle{unsrtnat}
\bibliography{references}  %%% Uncomment this line and comment out the ``thebibliography'' section below to use the external .bib file (using bibtex) .





% \end{thebibliography}

\end{document}